\documentclass[journal]{IEEEtran}

\usepackage{caption,subcaption,float}
\usepackage{graphicx}
\usepackage{multirow}
\usepackage{amsmath,fixmath,amsfonts}
\usepackage{array}
\usepackage{cite}
\usepackage[letterpaper]{geometry}
\usepackage{tikz}
\usepackage{enumitem}
\usepackage{tabularx}
\usepackage{array, multirow, bigdelim, makecell, booktabs} 
\usepackage{pgfplots}

\newcommand{\Lagr}{\mathcal{L}}
\usepackage{mwe} 

\begin{document}
\title{Tele-EvalNet: A Low-cost, Teleconsultation System for Home based Rehabilitation of Stroke Survivors using Multiscale CNN-LSTM Architecture}

\author{Aditya Kanade$^{1}$,~Mansi~Sharma$^{1}$,~\textit{Member},~\textit{IEEE},~M~Manivannan$^{2}$. 
\thanks{
$^1$The authors are with Department of Electrical Engineering, Indian Institute of Technology Madras, Chennai, India, 600036.

$^2$The author is with Department of Applied Mechanics, Indian Institute of Technology Madras, Chennai, India, 600036.

E-mail: \{ee20s086@smail,~mansisharma@ee\}.iitm.ac.in, mani@iitm.ac.in
}}

\maketitle

\begin{abstract}
Technology has an important role to play in the field of Rehabilitation, improving patient outcomes and reducing healthcare costs. However, existing approaches lack clinical validation, robustness and ease of use. We propose \emph{Tele-EvalNet}, a novel system consisting of two components: a \emph{live feedback model} and an \emph{overall performance evaluation model}. The live feedback model demonstrates feedback on exercise correctness with easy to understand instructions highlighted using color markers. The \emph{overall performance evaluation model} learns a mapping of joint data to scores, given to the performance by clinicians. The model does this by extracting clinically approved features from joint data. Further, these features are encoded to a lower dimensional space with an autoencoder. A novel multi-scale CNN-LSTM network is proposed to learn a mapping of performance data to the scores by leveraging features extracted at multiple scales. The proposed system shows a high degree of improvement in score predictions and outperforms the state-of-the-art rehabilitation models. The code and data will be made publicly available.
\end{abstract}

\IEEEpeerreviewmaketitle

\section{Introduction}
\label{sec:intro}
The number of stroke survivors are increasing in the world. A great effort is directed at improving the quality of life of stroke survivors with wide range of technology that has been developed towards this aim. However, very few medical institutions are exploiting computer based rehabilitation tools \cite{home-rehab}. Primary reason being the lack of clinical validation in the existing technological solutions. Literature surveys indicate that more than 90\% of rehabilitation sessions are carried out at homes \cite{ninty-pc-rehab}. The patients either have to self monitor or take help from the family members to monitor progress in the rehabilitation program. This voluntary nature of home-based rehabilitation program leads to low levels of patient adherence. This leads to prolonged post hospitalization recovery \cite{prolonged-1, prolonged-2}. Multiple tools and devices have been developed to support home based rehabilitation, such as the robotic assistive systems \cite{robotic-assistance}, virtual reality and gaming interfaces \cite{gaming-interface}, Kinect based assistants \cite{ninty-pc-rehab}. However, the current solutions lack robustness and clinical validity.

Liao et al. \cite{state-of-art} studied the problem of physical rehabilitation.
Their model uses UI-PRMD dataset \cite{ui-prmd}. The dataset consists of 10 healthy volunteers performing 10 physical rehabilitation exercises. The dataset is recorded on two systems, optical marker based Vicon Tracking system and a markerless Kinect system. They use Vicon tracking system data for their model. The authors use an autoencoder framework to compress the high dimensional joint data. Their model captures human variability with a mixture of gaussians. The model is trained using only correct performance data. Scores for all the performances are generated by taking negative log likelihood from the trained model, and are mapped between the range $[0, 1]$. Finally, they train a CNN model to map the full dimensional joint data to the generated scores. However, we found this approach not suitable when considering the Kinect data, due to high amount of noise. 

With our motivation stemming from the idea that goal based therapy interventions have better outcomes in a rehabilitation scenario \cite{goal-based-rehab}. We propose a novel system, which provides feedback to the patient on their live performance of exercise, with easy to understand instructions in color coding. Thus, enabling patients understand and correct their performance in the rehabilitation process. A novel CNN architecture is proposed, dubbed as \emph{Tele-EvalNet}. The \emph{Tele-EvalNet} is inspired from multi scale CNN architecture presented by Cui et al. \cite{mscnn}. We propose the use of clinically validated features extracted from the joint data to train an autoencoder, which learns a low dimensional representation of the features. We train our \emph{Tele-EvalNet} model using the low dimensional features. We show that use of context \cite{context} while feeding to \emph{Tele-EvalNet} improves performance of the network. The proposed \emph{Tele-EvalNet} achieves better performance compared to current state-of-the-art \cite{state-of-art}.

\section{Data Set}
The  KIMORE dataset is a free and open-source dataset \cite{kimore}. It has recordings of five rehabilitation exercises captured via a RGB-D camera. Table \ref{tab:exercise-kimore} lists out the exercises covered in our model.

\begin{table}[ht]
    \centering
    \caption{Exercises in KIMORE Dataset}
    \begin{tabular}{c c}
       \hline\hline
       Order  &  Exercise\\
       \hline
       E1  &  Lifting of Arms\\
       E2  &  Lateral tilt of the trunk with arms in extension\\
       E3  &  Trunk Rotation\\
       E4  &  Pelvic Rotation on the Transverse Plane\\
       E5  &  Squatting\\
       \hline \hline
    \end{tabular}
    \label{tab:exercise-kimore}
\end{table}
The data is available in three formats RGB-D, depth map, skeleton joint positions and skeleton joint orientations for every frame. The rehabilitation exercises are chosen for lower back pain. The authors have provided a set of meaningful features chosen by clinicians to describe each exercise. We have used this set of features to extract more meaningful data out of the raw skeletal data. Thus, using these features improve clinical validity of our system. The performance of both healthy subjects and subjects with motor dysfunction are evaluated on a scale of $[0, 50]$ by clinicians and are given as part of the dataset. KIMORE dataset considers a heterogeneous population to avoid sampling bias. Table \ref{population-dist} shows  distribution of the population.

\begin{table}[ht]
\caption{Population distribution in KIMORE dataset}
\renewcommand{\theadfont}{\normalsize\bfseries}
\begin{tabular}{c c cl}
\hline\hline
\thead{Population}  & \thead{Sample}\\
\cmidrule{1-3}
Healthy Subjects  &  44(29-M,15-W)\\
 Stroke Patients  &  10 & \hspace{-3em}\rdelim\}{4}{*}[(15-M,19-W)] \\
Parkinsons Patients  &  16 \\
Patients with backpain due \\to Spondylosis  &  8 \\
\hline\hline
\end{tabular}
\label{population-dist} 
\end{table}

There are a total of 78 participants, divided into 2 groups with 44 healthy subjects and 34 with motor dysfunction. Let the set $\textbf{E} = \{e_{p} | p \in \{1...N\}\}$ denote sensory data collected for a particular exercise, performed by 44 healthy participants, where $e_{p} = \{ e_{p}^{(1)}, e_{p}^{(2)}, ..., e_{p}^{(m)}\}$ represents the set of $m$ frames recorded for each individual and $e_p^{m} = \{e_p^{(m,1)}$, $e_p^{(m,2)}$, $..., e_p^{(m,k)}\}$ is the set of extracted features in each frame. The dimensionality of data for healthy individuals is $(44 \times M \times K)$. Similarly, let $\textbf{S} = \{s_{p} | p \in \{1...Q\}\}$ denotes the sensory data collected for a particular exercise performed by 10 stroke patients, where $s_{p} = \{ s_{p}^{(1)}, s_{p}^{(2)}, ..., s_{p}^{(m)}\}$ represents the set of $m$ frames recorded for each individual. The set $s_p^{m} = \{s_p^{(m,1)}, s_p^{(m,2)}, ..., s_p^{(m,k)}\}$ is of extracted features in each frame. The dimensionality of data for patients with stroke is $(10 \times M \times K)$. The Kinect outputs 3-D positions of 25 joints as well as their quaternion rotations with respect to the spine base. The feature vector size at every frame is 75 for 3-D joint positions and 100 for joint orientations. Feature vector size depends on the exercise. Please refer to \cite{kimore} for more details.

Note: The unequal number of frames between various subjects for a exercise are equalized by interpolating the mean of the number of frames among all subjects.

\section{Proposed Model}
\label{sec:majhead}
\subsection{Live Motion Feedback System}

\begin{figure}[h!]
  \centering
  \centerline{\includegraphics[width=6.5cm]{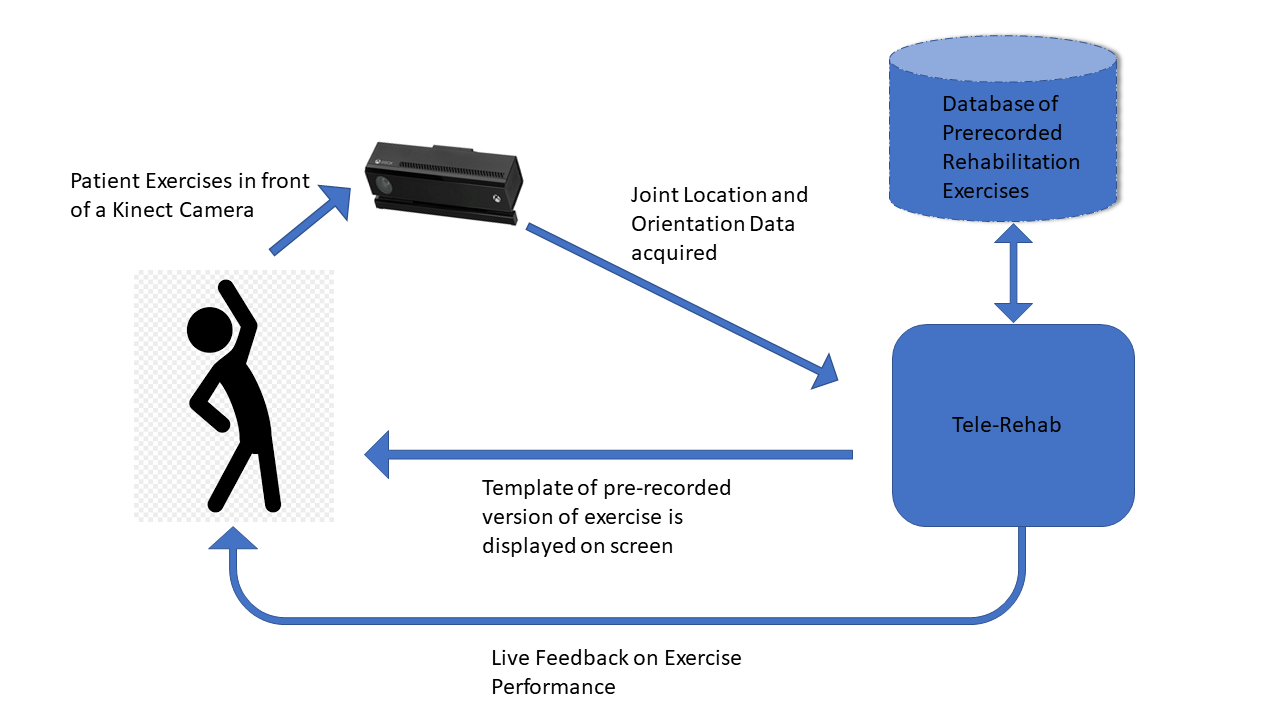}}
  \caption{Live Feedback System Architecture}
\label{fig:live-model-arch}
\end{figure}

The proposed system aims at guiding and monitoring Stroke patients without presence of a therapist. Fig. \ref{fig:live-model-arch} shows the architecture of Live Feedback System. We propose a color coded feedback system, where the patients can correct themselves while performing the exercise by observing easy to understand  feedback instructions. Feedback proposals are generated by comparison with the joint data of the template. The instructions are given based on the measure of dissimilarity of joint orientations between the patient and template. Dissimilarity is calculated for each joint at every frame as follows

\begin{equation}
T_{i} = \sum_{k=i}^{k=i+3}\|e_{p}^{(m, k)} - s_{p}^{(m, k)}\|
\end{equation}

\begin{figure}[h!]
  \centering
  \centerline{\includegraphics[width=6.5cm]{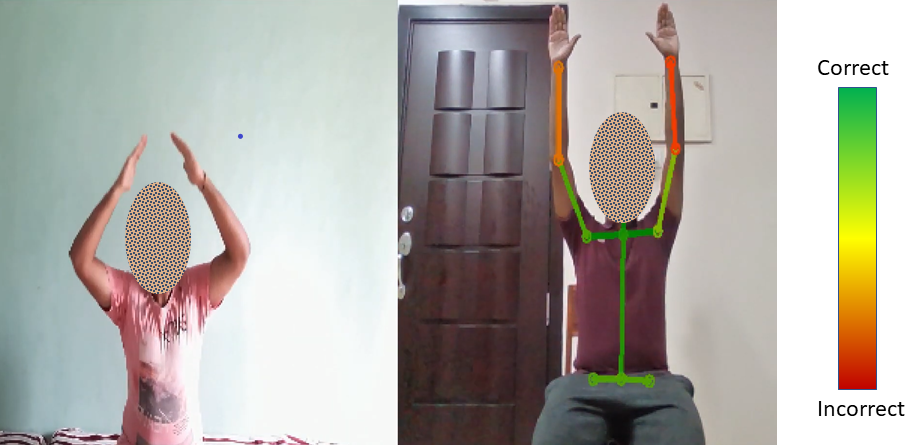}}
  \caption{Color Coded Feedback cues}
\label{fig:color-feedback}
\end{figure}

Joints are colored based on the value of $\emph{T}_{i}$ as it is a measure of dissimilarity. The coding scheme is linearly graded. As shown in Fig. \ref{fig:color-feedback}, the red indicates high degree of dissimilarity, yellow indicates mild degree of dissimilarity and green indicates high degree of similarity. Fig. \ref{fig:color-feedback} depicts monitor display that plays recording of the template video. Left half shows the recording of selected exercise and the right half shows live feed of the patient. The skeletal joints are overlaid on top of the patients body. Note that joints on torso are shown in green as they match the template. However, the upper limbs are colored in shades of red as they show high degree of dissimilarity with the templates motion. We would like to point out that, even if the patient performs the sequence correctly, but much slower than the template, they will see shades of red and green as feedback, due to the frame wise comparison done by the system. However, speed of performance will not affect the score predicted by the system.

\subsection{Overall Performance Evaluation}

\begin{figure}[h]
  \centering
  \centerline{\includegraphics[width=6.5cm]{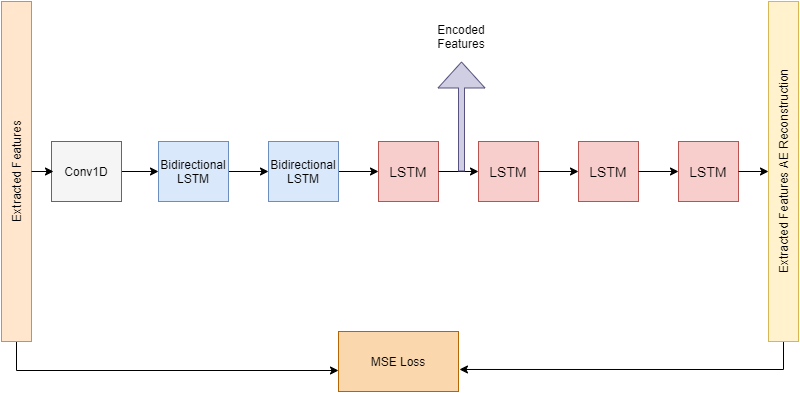}}
  \caption{Autoencoder Architecture}
\label{fig:autoencoder-arch}
\end{figure}

\subsubsection{Dimensionlity Reduction}
Dimensionality reduction techniques are used to get a stable and statistically accurate model using machine learning \cite{dim-reduce-better}. These techniques learn a lower dimensional representation of a high dimensional data by removing highly correlated dimensions and also learn structures in data that might not be apparent. 

One of the most common techniques of dimensionality reduction is the Principle Component Analysis (PCA). PCA decomposes a D dimensional data into it's eigen vectors and eigen values. It chooses only M eigen values, such that, M $<$ D, and projects the input data into space spanned by M eigen vectors. The technique works well for large class of problems, but it is a linear model and non-linearities in the data are not captured well \cite{pca-compare}. To overcome the challenge of modelling non-linear relationship, autoencoders were introduced \cite{deep-learning}. An autoencoder is defined by two components: 1) \emph{Encoder:} consists of a feedforward neural net and produces a latent representation, h = \emph{f}(x), 2) \emph{Decoder:} tries to reconstruct input using the latent representation, \^x = \emph{g}(h). The non-linear activations of the neural network helps capture non-linear structures in the data. The autoencoder is trained choosing reconstruction loss calculated as mean squared error between the input sample and the reconstructed sample from the decoder. 
A regularization factor $\Omega(\emph{f})$ is added to the reconstruction loss.
The regularization is performed using $L_1$ norm of the encoder weights. This assists the proposed model to generalize better \cite{regularization}.
\begin{equation}
    \Lagr_{total}(x) = \Lagr_{reconst.}(x;\emph{g}(\emph{f}(x))) + \lambda.\Omega(\emph{f})
\end{equation}
Fig. \ref{fig:autoencoder-arch} shows the architecture of the autoencoder. The encoder part consists of two Bidirectional LSTMs followed by a convolutional layer. The layers have increasingly lower feature vector representation. The latent vector, which represents the input sample in a lower dimensional space is collected from the output of the convolutional layer. This latent vector is passed on to the decoder part. The decoder has three stacked LSTMs which expand the feature dimension back to the input. We have chosen LSTMs as they help model timeseries data better.

\subsubsection{Tele-EvalNet}
\begin{figure}[h]
  \centering
  \centerline{\includegraphics[width=6.5cm]{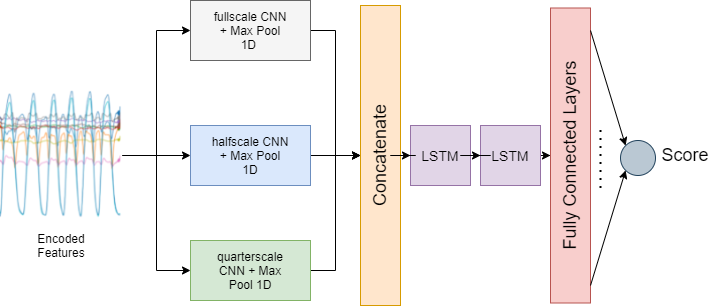}}
  \caption{CNN Architecture}
\label{fig:cnn-lstm}
\end{figure}

With the aim of capturing the scoring pattern of a clinician when evaluating performance of subjects. We propose \emph{Tele-EvalNet}, a novel CNN-LSTM based architecture which generates scores for the performed exercise. The model captures relationship between the movement data and the score given by clinician by looking at the joint data in multiple time scales. We were inspired by the work of Cui et al. \cite{mscnn} on time-series classification. In speech processing deep neural network models, it was seen that use of a context window by appending K past feature vectors and K future feature vectors improved performance of the network \cite{context}. Using this idea, we concatenate K past and future encoded joint data features and use this as input for the model. The architecture of the proposed model is shown in Fig. \ref{fig:cnn-lstm}. The input is fed to the proposed model in three time scales, 1) full-scale input, 2) input down-sampled by order of 2, 3) input down-sampled by order of 4. The model has three CNN branches, followed by a max-pooling layer. The features are collected from the max-pooling layer and combined using deep concatenation following the method of Szegedy et al. \cite{deep-concat}. These features are passed on to a stack of LSTMs which learn temporal relationships. Finally, the output of LSTM stack is passed on to a series of fully connected layers, which predict the final score.

\section{Experiments and Results}

\subsection{Results and Discussion}
In this section, we evaluate the performance of proposed Tele-EvalNet with current state-of-the-art. We have performed analysis in two section. Section \ref{dim-reduce} discusses the proposed models performance against current state-of-the-art methods for dimensionality reduction. Section \ref{rehab-net} discusses performance of the proposed model against current state-of-the-art for exercise evaluation. All the models are trained and tested using KIMORE dataset \cite{kimore}.

\subsubsection{Dimensionality Reduction}
\label{dim-reduce}
The autoencoder is trained on features extracted from the joint data of 44 healthy individuals with a 0.8/0.2 train-validation split. Early stopping is used to avoid overfitting the training data. A patience value of 1000 epochs is set to monitor the validation loss. The data for both 44 healthy subjects and the subjects with stroke is encoded using the trained model. Table \ref{table:dim-reduce} shows the mean squared error between the input and the reconstructed sample of various state-of-the-art techniques. We show the proposed method outperforms state-of-the-art techniques by a large degree.

\begin{table}[ht!]
\centering
\begin{tabular}{||c | c | c ||} 
 \hline
 Algorithm & E1 & E2\\
 \hline\hline
 PCA & 0.5744 & 0.5659\\ 
 \hline
 Deep Rehabilitation Framework \cite{state-of-art} & 0.4803 & 0.6221\\
 \hline
 Proposed & \textbf{0.1157} & \textbf{0.0507}\\
 \hline
\end{tabular}
\caption{Dimensionlity Reduction techniques Comparison}
\label{table:dim-reduce}
\end{table}

\subsubsection{Tele-EvalNet results and discussion}
\label{rehab-net}
The proposed \emph{Tele-EvalNet} uses encoded data from the autoencoder as the input and the score given by clinician as the output. The clinical scores in KIMORE dataset are in range of $[0, 50]$, these are scaled to a range $[0, 1]$ for training. The network is trained on a 0.8/0.2 train-validation split. Early stopping is used to avoid overfitting the training data. A patience value of 25 epochs is set to monitor the validation loss. The network is trained on binary cross entropy loss between predicted score and the actual clinical scores. Fig 5 shows the MSE on Exercise 3 (Trunk Rotation) data with various context windows. No frames are concatenated for context window size of one, for data with context window size three each feature vector is concatenated with one past feature vector and one future feature vector. The plot in Fig. 5 clearly shows reduction in MSE with use of context. For further experimentation, we select context window size as 3. Table \ref{table:regression} shows the mean squared error in score predictions between various state-of-the-art methods. We see the proposed model outperforms the current state-of-the-art methods.

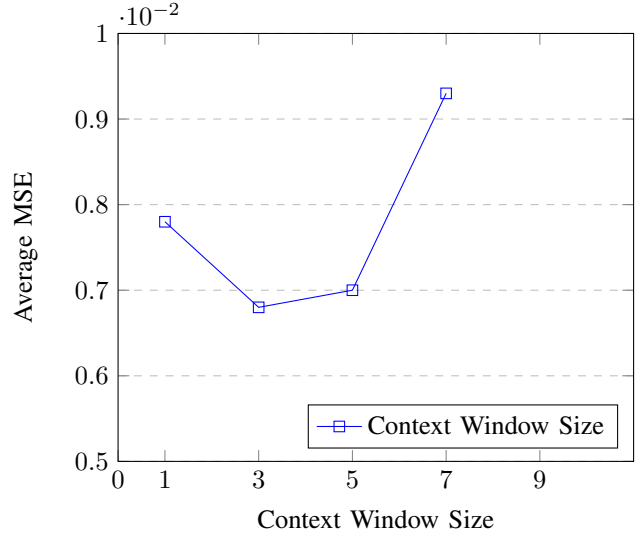
\begin{figure}[h]
\label{fig:context}
\centering
\begin{tikzpicture}
\begin{axis}[
    xlabel={Context Window Size},
    ylabel={Average MSE},
    xmin=0, xmax=11,
    ymin=0.005, ymax=0.01,
    xtick={0, 1, 3, 5, 7,  9},
    ytick={0.005, 0.006, 0.007,  0.008,0.009, 0.01},
    legend pos=south east,
    ymajorgrids=true,
    grid style=dashed,
]
\addplot[
    color=blue,
    mark=square,
    ]
    coordinates {
    (1, 0.0078)(3, 0.0068)(5, 0.0070)(7, 0.0093)
    };
    \legend{Context Window Size}
\end{axis}
\end{tikzpicture}
\caption{Comparison of average MSE with different window size}
\end{figure}

\begin{table}[h!]
\centering
\begin{tabular}{||c | c | c ||} 
 \hline
 Algorithm & E1 & E2\\ [0.5ex] 
 \hline\hline
 Deep LSTM & 0.0375 & 0.0491\\ 
 \hline
 Deep Rehabilitation Framework \cite{state-of-art} & 0.04975 & 0.03805\\ 
 \hline
 Proposed & \textbf{0.0224} & \textbf{0.0102}\\
 \hline
 Proposed w/o Encoding & 0.04081 & 0.03054\\
 \hline
\end{tabular}
\caption{Exercise Score Prediction Analysis}
\label{table:regression}
\end{table}

The results demonstrate better performance of the proposed model in predicting the scores as compared to \cite{state-of-art}. The improvement in score prediction is aided with a smaller size of model and number of parameters required to learn the mapping due to use of the extracted features.

\section{Conclusion}

This paper proposes a comprehensive framework for home based rehabilitation. The framework consists of two parts, 1) live feedback model, 2) overall performance evaluation model. The live feedback model gives easy to follow instructions to the patient, helping them correct exercise performance. The overall performance evaluation model extracts clinically validated features from joint data and trains an autoencoder to encode them to lower dimensions. \emph{Tele-EvalNet} is proposed, which is a novel CNN-LSTM based architecture that learns a mapping from the encoded features to clinical scores given to the exercise performance. The models are trained and evaluated on KIMORE dataset. We also show use of context for encoded joint data improves performance of \emph{Tele-Evalnet}. We show our models outperforming the current state-of-the-art methods \cite{state-of-art}.

\section{Future Work}
The current model does not support recording video of the patient due to privacy concerns, however, taking the patient into confidence over their videos being collected. A free-viewpoint video can be generated and sent to the clinician to help them evaluate the patients performance \cite{nerf, 3d-tv}. The monitor display being used in the proposed model can be replaced by 3D augmented reality glasses, this can be used to overlay charts, show scores to monitor performance and also improve portability of the model \cite{ar-app, rgb-face}.

\bibliographystyle{ieeetr}
\footnotesize
\bibliography{root.bib}
\end{document}